# ChatGPT v Bard v Bing v Claude 2 v Aria v human-expert. How good are AI chatbots at scientific writing?

Edisa Lozić[1] and Benjamin Štular [1,*]

[1] Research Centre of the Slovenian Academy of Sciences and Arts, 1000 Ljubljana, Slovenia; benjamin.stular@zrc-sazu.si

[*] Correspondence: benjamin.stular@zrc-sazu.si

**Abstract:** Historical emphasis on writing mastery has shifted with advances in generative AI, especially in scientific writing. This study analysed six AI chatbots for scholarly writing in humanities and archaeology. Using methods that assessed factual correctness and scientific contribution, ChatGPT-4 showed the highest quantitative accuracy, closely followed by ChatGPT-3.5, Bing, and Bard. However, Claude 2 and Aria scored considerably lower. Qualitatively, all AIs exhibited proficiency in merging existing knowledge, but none produced original scientific content. Interestingly, our findings suggest ChatGPT-4 might represent a plateau in large language model size. This research emphasizes the unique, intricate nature of human research, suggesting that AI's emulation of human originality in scientific writing is challenging. As of 2023, while AI has transformed content generation, it struggles with original contributions in humanities. This may change as AI chatbots continue to evolve into LLM-powered software.

## Highlights

- The article evaluates the scientific writing skills of six AI chatbots in the humanities and archaeology.

- The AI chatbots compared are: ChatGPT-3.5, ChatGPT-4, Bard, Bing Chatbot, Aria, and Claude 2. We also tested two ChatGPT-4 plugins: Bing and ScholarAI.

- ChatGPT-4 outperforms the other chatbots in quantitative accuracy, but is unable to "pass an undergraduate exam" in humanities.

- The study demonstrates the *limited potential of AI in generating original scientific contributions*, underscoring the unique value of human researchers.

- The growth in the size of large language models appears to have reached a plateau.

- As the size of language models like ChatGPT stabilises, it is important to understand their capabilities and limitations in the academic environment.

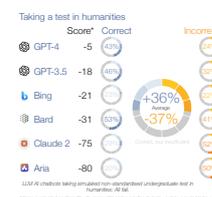

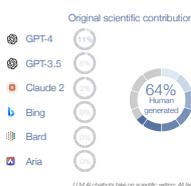

# Graphic extract

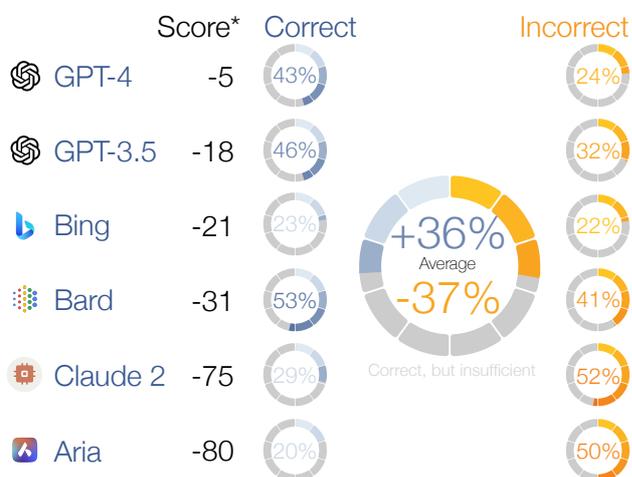

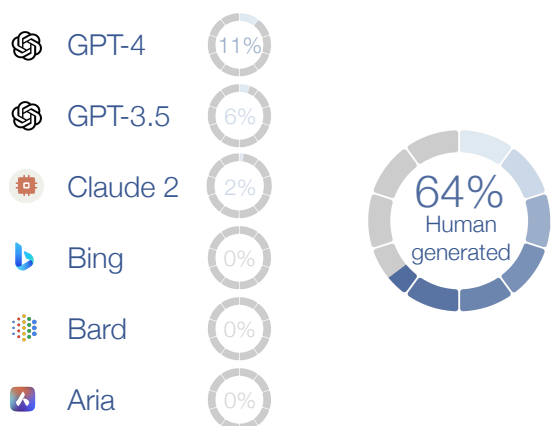

*LLM AI chatbots taking simulated non-standardised undergraduate test in humanities: All fail.*

*LLM AI chatbots take on scientific writing: All fail.*

\* Score is calculated as: Correct% - (2 x Incorrect%). Higher is better, students would be expected to have a positive score.

The race for parameters: LLMs grow exponentially, but after GPT-3 "jump" the content is only marginally improved.

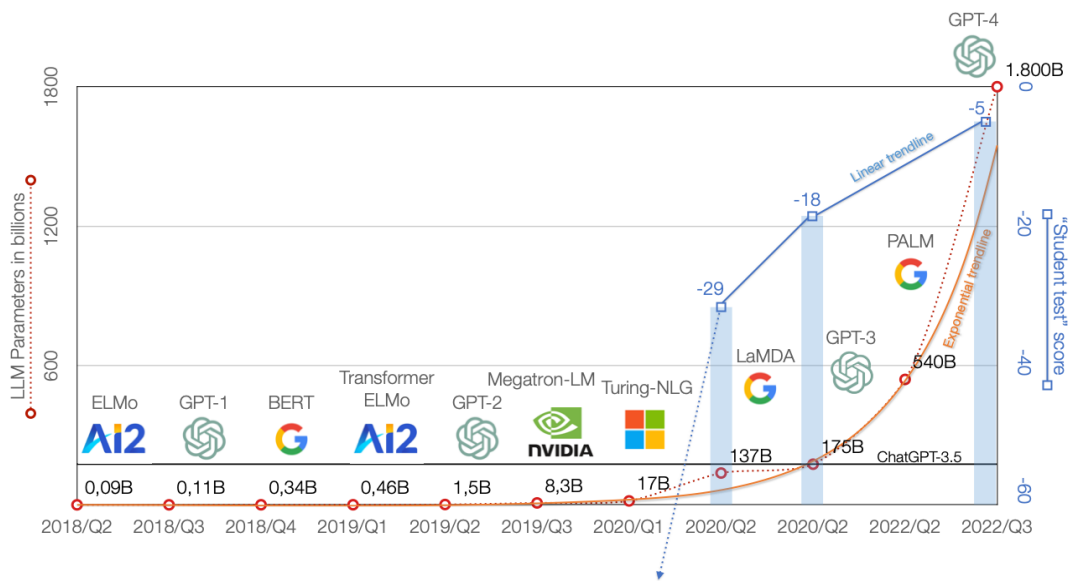

## 1. Introduction: AI's great inflection point

In recent human history, the ability to write well was considered essential to human progress and professionalism. Creative expression was traditionally seen as a defining characteristic of humanity and the pinnacle of human achievement. This view is still reflected in the way universities cultivate their students' writing skills. Until recently, novel artifacts such as literary works, scientific texts, art, and music were difficult to create and only attainable by talented experts [1–3].

We must reckon with that changing!

The current generation of openly available generative AI has rightly been called AI's great inflection point. Generative AI is shaping to become a general purpose technology [4], a "fundamental, horizontal technology that will touch everything in our lives" (Tim Cook, Apple CEO, speaking at Università Degli Studi di Napoli Federico II in Naples, Italy, 29.9.2022).

The most recent and disruptive advance in generative AI has been a leap-frog development in the field of large language models (hereafter LLMs). LLMs are based on deep neural networks and self-supervised learning that have been around for decades, but the amount of data that the current models were trained with lead to an unprecedented and, to some extent, unexpected performance leap. Current LLMs belong to foundation models that are pre-trained on a large datasets using self-supervision at scale and then adapted to a wide range of downstream tasks. This centralisation is crucial for harnessing the enormous computing power required to create them, but it also replicates all potential problems such as security risks and biases [3,5].

Currently, the most powerful LLMs are generative pretrained transformers (hereafter GPTs), which are based on the Transformer, a type of neural network architecture. The Transformer uses a mechanism called attention to weigh the influence of different input words on each output word. As a result, instead of processing words in a sentence sequentially, it constructs relationships between all words in a sentence at once [6]. Additional key advantage of GPTs over earlier models is that the learning process can be parallelised and the models can be trained on an unprecedented scale.

The scale of an LLM depends on the size of ingested datasets, the amount of training compute, and the number of parameters it can support [7,8]. Parameters are numerical values that determine how a neural network processes and generates natural language. The more parameters a model has, the more data it can learn from and the more complex tasks it can perform. GPT-3 from 2020, for example, supports 175 billion parameters and has been trained on 45 TB of text data, including almost the entire public web [9]. PaLM, the 2022 LLM from Google Research, is a 540-billion-parameter GPT model trained with the Pathways system [10], and GPT-4 launched in 2023 supports an estimated 1.8 trillion parameters [11].

That is 1,800,000,000,000 parameters with which the model interacts to generate each individual token (a word or a part of a word). Multiplied by ChatGPT's 100,000,000 monthly users each processing just one prompt with 100 tokens daily brings us to a staggering 18,000,000,000,000,000,000,000 or $18 * 10^{21}$ computations, which explains the daily costs of running ChatGPT at 700,000 $ [12]. One can only imagine the environmental costs of the operation.

When given an input or prompt, GPT LLMs are able to predict the next word in the context of all previous content and can thus generate creative outputs such as complete sentences and answers or even essays and poems. In essence, they generate a pattern of words based on the word patterns they have been trained by applying attention to context and a controlled amount of randomness. But because they have been trained on such a large amount of text, the quality of the text is such that GPT-4, for example, has been able to pass or even ace some standardised academic and professional tests [13].

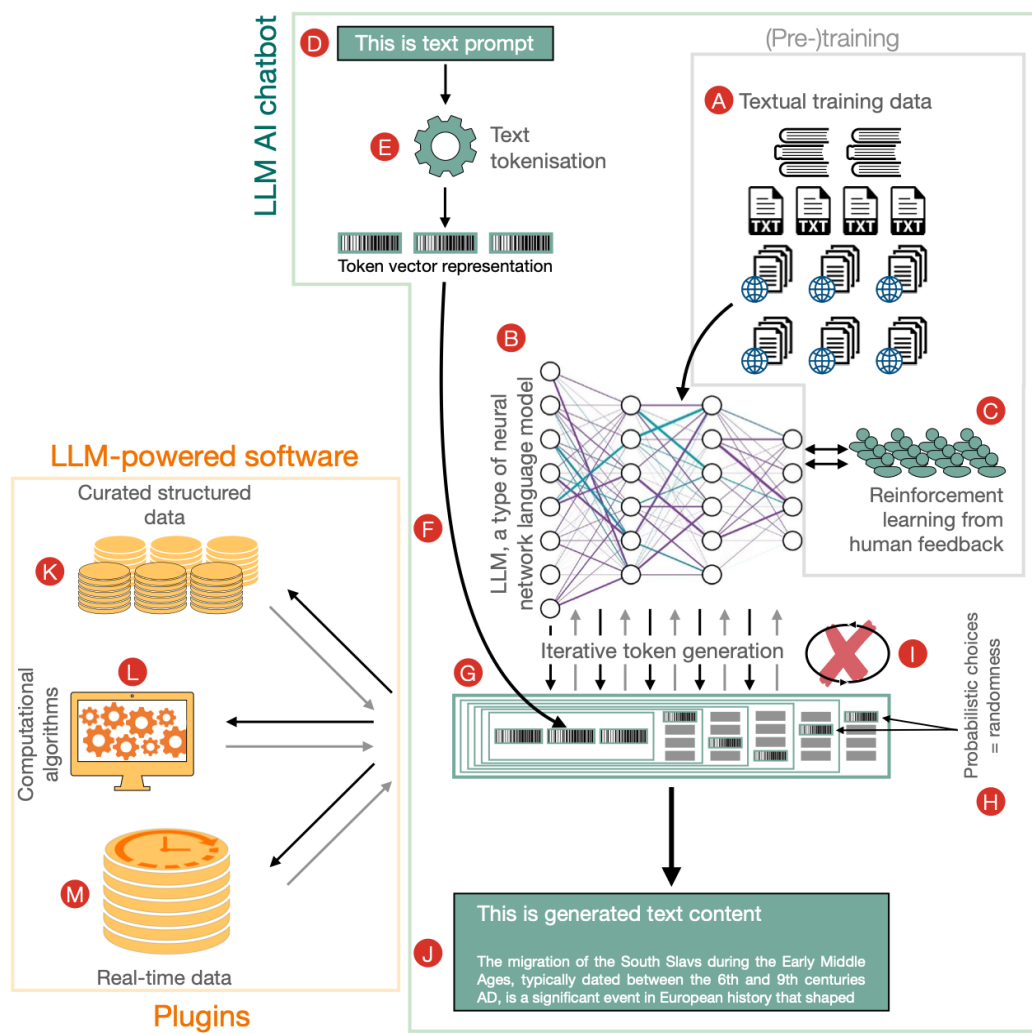

Fig. 1. A high level overview of an AI chatbot and an LLM-powered software (constructed based on information from [14,15]).

Users interact with a LLM via AI conversational agents or AI chatbots. An AI chatbot is a software application that uses AI techniques to simulate human conversation. In the past, AI chatbots did not use LLMs and in the future, AI chatbots are likely to use other resources besides LLMs, such as curated structured data, computational algorithms, and real-time data. Currently, however, AI chatbots are mostly used as conduits to GPT LLMs and thus the terms AI chatbot and LLM are sometimes used interchangeably (Fig. 1).

The most talked about AI chatbot in history was ChatGPT, built on the GPT LLM called GPT-3.5 [16] (https://chat.openai.com). It was released to the public in November 2022 and had 1 million active users in its first week (https://twitter.com/sama/status/1599668808285028353) and over 100 million after two months [17], comfortably beating the likes of Facebook, TikTok, Instagram and Uber to that milestone. Since then, Google has launched Bard [18] (https://bard.google.com), Microsoft the Bing chatbot [19](https://www.bing.com), OpenAI ChatGPT Plus [20] (https://chat.openai.com), Opera Aria [21] (https://www.opera.com/features/browser-ai), Anthropic Claude 2 [22,23] (https://claude.ai), and more.

AI chatbots have implications for a range of practices and disciplines, and for many facets of our daily lives. They can be used to compose emails, essays, computer code, and speeches or translate text into another language or simply give it a different tone. They can be empowering because they lower the barrier to entry, seamlessly complement human work, and make us more productive and creative [3,24,25]. Many (but mostly proponents of entities with a stake in AI chatbots) tout that AI chatbots will take the drudgery out of everyday office work by automating various tasks and ultimately increase productivity across the economy, e.g., [26,27].

But AI chatbots can also be terrifying and are seen by many as potentially threatening and having far-reaching negative consequences. They could reinforce the biases we already experience, undermine our trust in information, and take away our ability to determine what is real and what is not. Equally important, they are likely to upend the creative and knowledge industries and many types of work that have been primarily done by well-compensated professionals, such as artists, writers, executives, and programmers [3,24,25].

A recent study of the potential impact on the labour market found that LLMs and LLM-powered software will affect between 47% and 56% of all work-related tasks, with higher-income jobs being more affected. Of particular interest to our article is that roles that rely heavily on science and critical thinking show a negative correlation with exposure to LLMs, yet there is some exposure [4].

Thus, AI experts, journalists, policymakers, and the public are increasingly discussing a wide range of important and urgent risks of AI, such as AI race, organizational risks, rogue AIs, the reinforcement of social inequalities, remaking labour and expertise, and the exacerbation of environmental injustices [28–33]. Questions are being asked over safety [34,35], capabilities [36], massive workforce redundancy [37], and legality [38,39]. This has led to calls for a pause in AI development [40], although there are doubts that such attempts would have any impact [41].

In short, the opportunities that generative AI presents for our lives, our communities and our society are as great as the risks it poses [42]. There are those who believe that generative AI will change our work and our lives in general for the better. And there are others who believe that generative AI will disastrously encroach on areas best navigated by sentient beings. Regardless, all agree on an urgent need for safeguards [43–46] and the first steps have already been taken [47].

All this is even more true for the planned next step, artificial general intelligence (hereafter AGI). The term AGI describes AI systems that will generally be more intelligent than humans and are planned to become a reality within the next 10 years. After the introduction of the first AGI, the world could be extremely different from how it is today [48].

And how does academia and the way we create and write research and scholarly articles fit in? AI chatbots have the potential to revolutionise academia and scholarly publishing [49]. In fact, it seems that academia will be among the first industries to go through this process, since academics and students represented two of the top three occupational groups among the early adopters of ChatGPT [50].

AI chatbots—most of the attention to date was directed to ChatGPT—have already been recognised as a powerful tool for scientific writing that can help organise material, proofread, draft and generate summaries [51–53]. The scientific community is also actively testing their ability to generate entire papers with minimal human input. The consensus is that AI chatbots are able to create scientific essays and reports of scientific experiments that appear credible but are a combination of true and entirely fabricated information [49,51,54–63].

Unfortunately, there is a public perception that ChatGPT is already capable of generating academic papers that get peer-reviewed and published, e.g., [64,65], which may add to public scepticism about science. This is not the case. For example, the ArXiv repository of pre-prints (https://arxiv.org), the AI community's most popular publication forum, shows no results for ChatGPT (in any variation) as a (co-)author (tested on 16 August 2023). We are aware of a single such attempt that has attracted a lot of public attention, but did not get peer reviewed and has no notable scientific value [55].

Regardless, there is a clear consensus among researchers that AI chatbots will be widely adopted in scientific writing in the near future, and it is thus crucial to reach an accord on how to regulate their use [63,66].

However, to successfully discuss regulation, a better understanding of AI chatbots is needed. This requires more systematic testing of their capabilities, which will provide a more robust understanding of the strengths and weaknesses of the technology. This process has been likened to the processes drugs go through to gain approval. Assessment of AI systems could allow them to be deemed safe for certain applications and explain to users where they might fail [66].

To this end, there is a growing body of testing and benchmarking of generative AI models, e.g., [13,67,68]. The standard methodology in machine learning is to evaluate the system against a set of standard benchmark datasets, ensuring that these are independent of the training data and span a range of tasks and domains. This strategy aims to distinguish real learning from mere memorisation. However, this approach is not ideally suited to our needs and to the study of LLM-based AI chatbots for three reasons. First, only the creators of proprietary LLMs have access to all the training details needed for detailed benchmark results. Second, one of the key aspects of the intelligence of LLMs is their generality and ability to perform tasks that go beyond the typical scope of narrow AI systems. Metric of evaluation benchmarks designed for such generative or interactive tasks remain a challenge. The third and perhaps most important reason is that in this article we are interested in how well AI chatbots perform in human tasks. To evaluate this, methods closer to traditional psychology leveraging human creativity and curiosity are needed [69].

Such an approach has already been taken, and there are several evaluations of the performance of AI chatbots in scientific writing, but most of them focus on medicine and similar fields [49,51,56–63,70–72]. We are not aware of any such test designed specifically for humanities. Therefore, more tests and, we believe, more types of tests on the performance of AI chatbots in scientific writing are urgently needed.

With this in mind, the aim of this article was to design and conduct a test of AI chatbots' abilities in scientific writing in the humanities. First, we were interested in their ability to generate correct answers to complex scientific questions. Second, we tested their capacity to generate original scientific contributions in humanities research. Since AI chatbots are developing at a staggering pace, our results apply to the state of affairs in the third quarter od 2023 (23Q3).

To achieve this, we used an interdisciplinary case study combining archaeology, historiography, linguistics and genetic history. We created two one-shot prompts to ask complex scientific questions and we fed them to each of the six AI chatbots tested: ChatGPT-3.5, ChatGPT-4, Bard, Bing Chatbot, Aria, and Claude 2. We also tested two ChatGPT-4 plugins: Bing and ScholarAI. The generated content was evaluated against each other and against the human-generated content.

The main reason for the comparison between the AI chatbots was to create a baseline for the staggering speed at which they evolve. Comparison with human-generated responses served as a starting point for the discussion on how upending, transformative, and disruptive generative AI models will be in the humanities in the future.

## 2. Materials and methods

*2.1 AI Chatbots*

This section describes the AI chatbots that were tested. As they are all proprietary commercial products, there is often not much detail available, let alone in the form of peer-reviewed articles. Our descriptions are therefore based on various sources such as blogs, social media and help pages. We do not go into the specifics of the underlying LLMs, as this is a specialised topic on which there is an extensive literature, e.g., [73,74].

First, the criteria for selecting the six AI chatbots should be elucidated. As mentioned earlier, most of the previous studies have only analysed ChatGPT-3.5. Its inclusion, as well as that of its successor ChatGPT-4, was therefore a given. The Bing chatbot was included because it was arguably the most advanced freely available AI chatbot at the time of the test. Bard was included because it is seen by many as the only challenger to ChatGPT's hegemony. We also wanted to include two chatbots that use an application programming interface (hereafter API) to access LLM. APIs are the only available means for "smaller" developers, i.e., anyone other than OpenAI/Microsoft or Google, to access state-of-the-art LLMs. We chose Aria and Claude 2, which use APIs from OpenAI and Google, respectively. If Aria and Claude 2 performed on par with ChatGPT and Bard, it would signal that generative AI technology is indeed being developed openly "for all humanity", and vice versa. The two plugins, ChatGPT with Bing and ScholarAI, were chosen from the rapidly growing selection as the two most relevant to the task of scientific writing. Baidu's ERNIE bot (https://yiyan.baidu.com), on the other hand, was not considered because at the time, it was only available with a Chinese interface and required a Baidu login and the Baidu app (also only available in Chinese) to use.

*ChatGPT-3.5*, sometimes called ChatGPT or GPT-3.5, is an AI chatbot offered as a free service by OpenAI (https://chat.openai.com; accessed on 11 October 2023). It was fine-tuned from a model in the GPT-3.5 series, more specifically gpt-3.5-turbo, in 2022. This autoregressive language model has the same number of parameters as the largest model from the 2020 GPT-3 series, namely 175 billion. The model was trained using the same methods as InstructGPT-3, but with slight differences in the data collection setup and by using supervised fine-tuning. To predict the next token in the text, it was pre-trained on approximately half a trillion words and improved by task-specific fine-tuning datasets with thousands or tens of thousands of examples that primarily used reinforcement learning from human feedback [75]. ChatGPT-3.5 achieved strong performance on many NLP datasets, including translation and question answering, as well as on several tasks requiring on-the-fly reasoning or domain adaptations. Its breakthrough results paved the way for a new generation of LLMs by demonstrating that scaling language models exponentially increases performance. GPT-3.5 is also available as an API [9]. It may come as a surprise that the core team that developed ChatGPT initially consisted of only approximately 100+ experts, although crowdworkers were also involved as a so-called Short term alignment team [76].

*ChatGPT-4*, also known as ChatGPT Plus or GPT-4, is an evolution of ChatGPT-3.5. It is based on a new model with 10 times the number of parameters estimated at 1.8 trillion [11]. The chatbot is available as a paid service ChatGPT Plus (https://chat.openai.com) and it is based on a large-scale multimodal model that can accept both image and text input and produce text output, but at the time of our test (August 2023) only text input was possible. Another core component compared to ChatGPT-3.5 was the development of an infrastructure and optimisation methods that behave predictably across a wide range of scales. Reinforcement learning from human feedback is also an ongoing process. ChatGPT-4 demonstrates human-level performance on various standardised professional and academic benchmarks, including passing a simulated bar exam with a score near the top 10% of test takers [13].

The ChatGPT service has opened up to third-party plugins [77] and we tested two relevant plugins that existed in July 2023. The plugins are available within the ChatGPT Plus service, but the whole service is in beta testing.

ChatGPT-4 with Bing plugin (henceforth *ChatGPT w/Bing*) has no available information on how it processes its prompts. Furthermore, the plugin was only briefly available as an early Beta in late June, it was temporarily unavailable in July and was missing from the plugin store in August. The results must therefore be considered only as an indicator of the final capabilities.

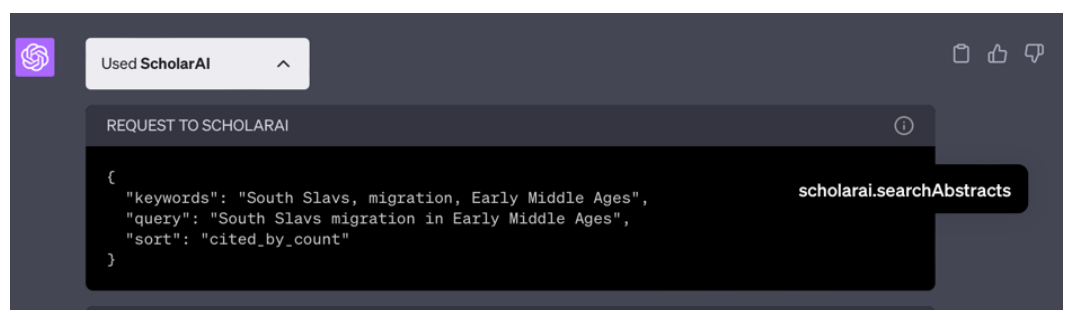

Fig. 2. Screengrab of the ScholarAI plugin for ChatGPT plus demonstrating its inner workings.

ChatGPT-4 with ScholarAI plugin (henceforth ScholarAI) was also under active development during our test. No documentation was available, but the plugin provided metadata about how it processed the prompt (Appendix A: L. 588, footnote 24; Fig. 2) and ScholarAI provided additional information (personal communication with Lakshya Bakshi, CTO and Co-founder of ScholarAI). First, it extracted keywords from the prompt, which were then recombined into a query. Based on this query, it returned the top results from the ScholarAI database of "40M+ peer-reviewed papers". The user then either confirmed the selection or requested further search. When the user was satisfied with the selection of articles, ScholarAI fed the ChatGPT-4 LLM with content. ScholarAI ensures that the LLM receives the source data and tries to get ChatGPT to discuss only the content provided to it, but this is a work in progress.

The *Bing Chatbot* is available either as a Bing Chat or a Bing Compose service and can be used free of charge in the Microsoft Edge browser and in the Bing app (https://www.bing.com/new). The Bing Chatbot is based on proprietary technology called Prometheus, an AI model that combines Bing's search engine with ChatGPT-4. When prompted by a user, it iteratively generates a series of internal queries through a component called Bing Orchestrator. By selecting the relevant internal queries and leveraging the corresponding Bing search results, the model receives up-to-date information so that it can answer topical questions and reduce inaccuracies. For each search, it ingests about 128,000 words from Bing results before generating a response for the user. In the final step, Prometheus adds relevant Bing search responses and is also able to integrate citations into the generated content. This is how Prometheus grounds ChatGPT-4. However, for prompts that the system considers to be simple, it generates responses using

Microsoft's Turing language models, which consume less computing power (https://www.linkedin.com/pulse/building-new-bing-jordi-ribas; Fig. 3). The most obvious difference between Chat and Compose is that the former is limited to about 200 words and the latter offers more setting options for the generated content. In our tests, however, the two generated significantly different content (Appendix A).

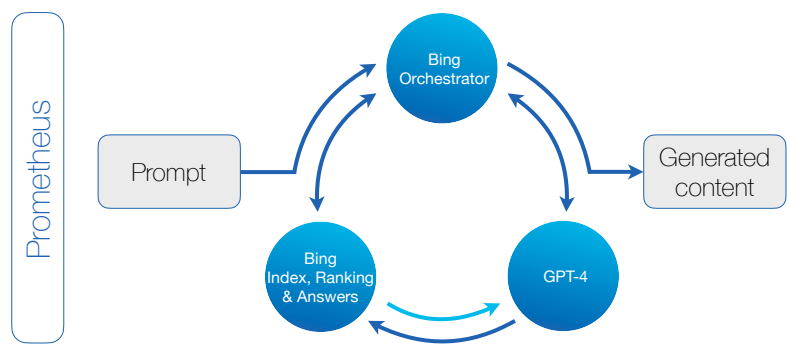

Fig. 3. Prometheus, a high level overview of the workflow (adapted after https://www.linkedin.com/pulse/building-new-bing-jordi-ribas).

*Aria* is a browser AI available for free in the Opera browser (https://www.opera.com). It was "built on Opera's AI engine, based on several Large Language Models (e.g. Generative Pre-trained Transformer built by OpenAI)." It was primarily designed to support and enhance the browsing experience by compiling information from various online sources. In addition, it can be used as an AI chatbot to assist in generating texts on any topic in any style for which it can ingest web content. One of Aria's most important features is its ability to convert certain keywords in its responses into web links [21,78].

*Bard* is an AI chatbot available as a free service from Google to a (rapidly growing) limited number of users [18]. It is based on a lightweight and optimised version of Language Models for Dialogue Applications (hereafter LaMDA). LaMDA is a family of GPT LLMs [6] designed specifically for dialogues. First announced during the 2021 Google I/O keynote, it has up to 137 billion parameters and has been pre-trained with 1.56 trillion words from public dialogue data and web text. In addition, LaMDA was fine-tunned for safety and factual grounding. The former was achieved by filtering candidate responses using a LaMDA classifier and data annotated by crowdworkers, the latter by enabling the model to consult external sources, such as an information retrieval system, a language translator, and a calculator. Therefore, Bard interacts with an external information retrieval system to improve the accuracy of facts provided to the user [79–81].

*Claude 2* is an AI chatbot released by Anthropic, a public benefit corporation, and is currently freely available in the US and UK via a website (claude.ai) and as an API. The underlying LLM was developed on Google's hardware, but few details are available about the model's pre-training. Claude 2 stands out for its fine-tuning. In addition to reinforcement learning with human feedback, so-called constitutional AI was used. This method was developed by Anthropic researchers and requires the model to respond to a large number of questions. It is then instructed to make its answers less harmful by adhering to the "constitution". Finally, the model is adjusted to generate such less harmful content in response to the initial prompt. Claude's constitution, a document created by the developers, is based on the United Nations Declaration of Human Rights and other principles that reflect non-Western perspectives. Another unique feature of the Claude 2 AI chatbot is the size of its context window of 100,000 tokens, or about 75,000 words, which allows users to paste or upload large documents or entire books and prompt questions based on their content [22,82,83].

In summary, in this article we compare three different types of AI tools: AI chatbots (ChatGPT, Bard, ChatGPT w/ Bing, Claude 2), a new generation of search engines (Bing

Chat, Aria), and tools powered by ChatGPT (Bing Compose, ScholarAI). Despite these differences, all except Bing Chat were able to comply with our prompts and were therefore suitable for our test.

It must be emphasized that the tested AI chatbots were designed as general purpose AI chatbots capable of human-like conversation, and not to "do science". Such downstream applications can be expected in the near future.

*2.2 Domain-specific scientific background*

The case study chosen for testing AI chatbots was the migration of the South Slavs with a follow-up prompt on the Alpine Slavs, a subgroup of the South Slavs. The authors' thematic explanation of the case study can be found in the appendix (Appendix A: L. 383-439 and 799-855).

The migration of the Slavs, including the South Slavs, has been a research topic for almost a century. Notwithstanding this, the rapid spread of the Slavic language in the second half of the first millennium CE remains a controversial topic [84–92]. It is part of the grand narrative of the "dawn of European civilisation". The Europe we live in today emerged in the centuries after the decline of the Roman Empire and was importantly shaped by the ancient Slavs, among others.

The current scientific debate on this issue revolves around the gene pool landscape on the one hand and the so-called ethnic landscape on the other. Until the 1950s, migration was assumed to be the main process of change, e.g., [93] and peoples and tribes were understood as caroming around the continent like culture-bearing billiard balls [94]. It was during this period that the term Migration period was coined. Since the 1960s, the understanding of ethnic identity has shifted to the concept of dispersed identities, which states that people fluidly adopt different identities as changing social circumstances dictate, e.g., [95]. Today, most assume that hardly any physical migration has taken place, but rather that ideas and knowledge have been passed on, e.g., [96]. However, recent research in the field of DNA, ancient DNA, and deep data analysis supported by machine learning is providing increasingly compelling evidence that, at least in the case of the South Slavs, physical migrations of people and peoples took place [92].

*2.3 Text prompts*

Our experiment was based on asking generative AI models two specific scientific questions. We designed two text prompts that were precise enough to produce the desired result without follow-up prompts.

The selected case study spans several academic fields, one of which can be considered a natural science (DNA analysis), one a humanities science (historiography) and two a humanities science with links to natural science (archaeology, linguistics). In the USA, archaeology is also considered a social science in certain contexts.

The two text prompts were:
- Q1: What is scientific explanation for migration of South Slavs in Early Middle Ages? Write 500 words using formal language and provide references where possible.
- Q2: What is scientific explanation for the settlement of Alpine Slavs in Early Middle Ages? Write 500 words using formal language and provide references where possible.

*Q1.* The first prompt is a complex scientific question on the subject of Early Medieval studies. To discuss it requires knowledge of archaeology, historiography, linguistics, and DNA studies. However, the topic is relatively broad. Spatially, it covers an entire European region, the Balkans. Its scientific background, the migration of the Slavs, is relevant to over 200 million modern Europeans. In short, although it is not one of the foremost topics in the Humanities or even for Early Medieval scholars, there are numerous researchers working on this topic and dozens of relevant scientific papers are published every year.

*Q2.* At first glance, the second prompt is almost exactly the same as the first, except for the target group, the Alpine Slavs. However, the added complexity of this prompt comes from the fact that it addresses a very narrow and specific topic. In fact, the only scholarly content on this topic is either more than half a century old, e.g., [97] and not available online, or it comes from a 2022 paper [92], which is too recent to be included in the datasets used for training ChatGPTs.

However, the key term "Alpine Slavs" is very specific. In response to the search term "settlement of the Alpine Slavs", the search engines Bing, Google and DuckDuckGo as well as Google Scholar return the mentioned article as the top hit after Wikipedia or the Encyclopaedia Britannica.

We therefore expected ChatGPT to respond well to Q1 but have more problems with Q2. On the other hand, AI chatbots with access to the current online content (Bing chatbot, GPT w/Bing) were expected to generate a high quality content for Q2 by sourcing it directly from the relevant article.

Our scientific questions are therefore so-called one-shot prompts, where the user provides the AI chatbot a single example of the desired task and then asks it to perform a similar task. It is well known that GPT LLMs are "few-shot learners" (BrownEt2020), i.e. they are much better at generating content when given more examples of the expected content. When using one-shot prompts, multiple refinement prompts are expected to improve results, e.g., [98].

However, few-shot prompts were not suitable for our testing purpose because they did not mimic a scientific question and a series of prompts would reduce the value of a direct comparison between different AI chatbots and introduce subjectivity. Therefore, our one-shot prompts were optimised for comparison rather than for generating the best possible content.

*2.4 Tagging and analysis*

There are several existing studies similar to ours, but they refer to other fields of science. Regardless, a brief overview of the methods used is in order. Altmäe and colleagues [51], for example, provided prompts and content generated by ChatGPT and then discussed the quality of the content. Petiška [56] focused only on references and analysed factors such as the number of citations, the date of publication, and the journal in which the paper was published. Májovský and colleagues [57] posed questions and prompts to the model and refined them iteratively to produce a complete article, which was then reviewed by relevant experts for accuracy and coherence. Buholayka and colleagues [59] tasked ChatGPT with writing a case report based on a draft report and evaluated its performance by comparing it to a case report written by human experts.

Our approach is similar to that of Májovský and colleagues [57], but there are three significant differences. First, we did not use iterative refinement to ensure comparability between different AI chatbots. Second, we did not generate a complete paper. Third, our review was both qualitative and quantitative, not just qualitative. This was achieved by tagging the content. The aim was to provide what is, to our knowledge, the first quantitative comparison of different AI chatbots on the subject of scientific writing.

The content generated by each of the tested AI chatbots was tagged for quantitative accuracy and qualitative precision.

*Quantitative accuracy* describes how accurate the content is in the opinion of the human experts (the authors). It was gradated into five classes:
- Correct: Factually correct and on par with the content created by human experts.
- Inadequate: Factually correct but falls short of the content created by human experts.
- Unverifiable: The statement cannot be verified or there is no expert consensus.

- w/ Errors: Mostly factually correct, but with important errors that change the meaning.
- Incorrect: Factually incorrect.

Quantitative accuracy is the measurement most commonly applied to AI-generated content, providing a quantifiable index of how trustworthy the tested AI chatbot is for the task at hand. From the perspective of academia, it can be understood as similar to the grading of a student's work. The questions in this case study are based on the grading of a senior undergraduate student attending a class on Early Medieval Archaeology.

*Qualitative precision* describes how "good" the content is in the opinion of a human expert. In other words, how it compares to a human-generated response in the context of scientific writing. It was gradated into four classes:

- Original scientific contribution.
- Derivative scientific contribution.
- Generic content not directly related to the question.
- Incorrect: Factually incorrect, containing errors that change the meaning, or disputed (the last three classes of above quantitative tagging combined).

Qualitative precision, as we have defined it, is specific to testing AI-generated content for the purpose of scientific writing and, to our knowledge, has not yet been used. The reason for the insufficient development of such indices is mainly that the current generation of AI chatbots is not expected to generate original scientific content. However, in the near future AGI will be expected to produce original scientific content. The qualitative precision index was therefore developed with an eye to the future as a measure of how close the tested AI chatbots are to AGI.

From an academic perspective, qualitative precision can be understood in a similar way to peer review of a scientific paper. As with any peer review, e.g., [99], it is a combination of objective and subjective evaluation.

It should be mentioned in passing that in the humanities an article usually consists of both derivative and original content. Typically, introduction and method are predominantly derivative, while results, discussion, and conclusion are predominantly original. The ratio of derivative to original content varies widely, depending on the discipline, topic, type of article, etc. Thus, the expected "perfect score" is not 100%, but in the order of 50+% of the original scientific contribution. To establish the baseline for our case study, we have tagged the responses generated by human experts (see Results section).

Both quantitative accuracy and qualitative precision tagging were performed by the two co-authors. To mimic the standard process used in grading students or reviewing scholarly articles we each made our own assessment and consulted to arrive at a unanimous decision. The results are shown in the appendices (qualitative accuracy: Appendix A; quantitative accuracy: Appendix B). Both co-authors have experience with both tasks in a professional capacity and both are experts on the topic, e.g., [92,100,101].

The tagged content was quantified and the results are discussed in the next section. Given the small amount of data generated by tagging, the observational method of analysis was sufficient. To sort the different AI chatbots according to the result, we calculated the accuracy score using the following formula: *Correct% - (2 x Incorrect%)*. The higher the score, the better. Students would need a positive score for a passing grade.

As the amount of tagging data increases in future projects, more sophisticated statistical methods will be used.

*2.5 Limitations of the method*

There are two limitations to our method. First, the case study is limited to a single (interdisciplinary) field in humanities and cannot address the differences between, for example, philosophy and geography. Second, only a limited number of human experts were involved.

For better results, we plan to extend this study to a series of field-specific studies and involve more human experts. However, in the current structure of public science, it takes years to accomplish such an organisational feat. Therefore, this article can be understood as an interim measure taken in response to the incredible speed at which AI chatbots are developing.

**3. Results**

*3.1 Quantitative accuracy: AI chatbots taking an undergraduate test*

The quantitative accuracy tagging was intended to objectively determine how correct the answers generated by the AI chatbots were (Fig. 4; Appendix A).

The highest accuracy score was achieved by ChatGPT-4 which also generated the highest percentage of correct content. On average, about half of the content provided was correct and about 1/5 was incorrect, with errors or unverifiable. However, as expected (see section 2.3), all of the incorrect content belonged to Q2 for which it could not source the relevant content from the 2022 article. Considering the complexity of the questions, the results were impressive, but far below what would be expected of, for example, a senior undergraduate student.

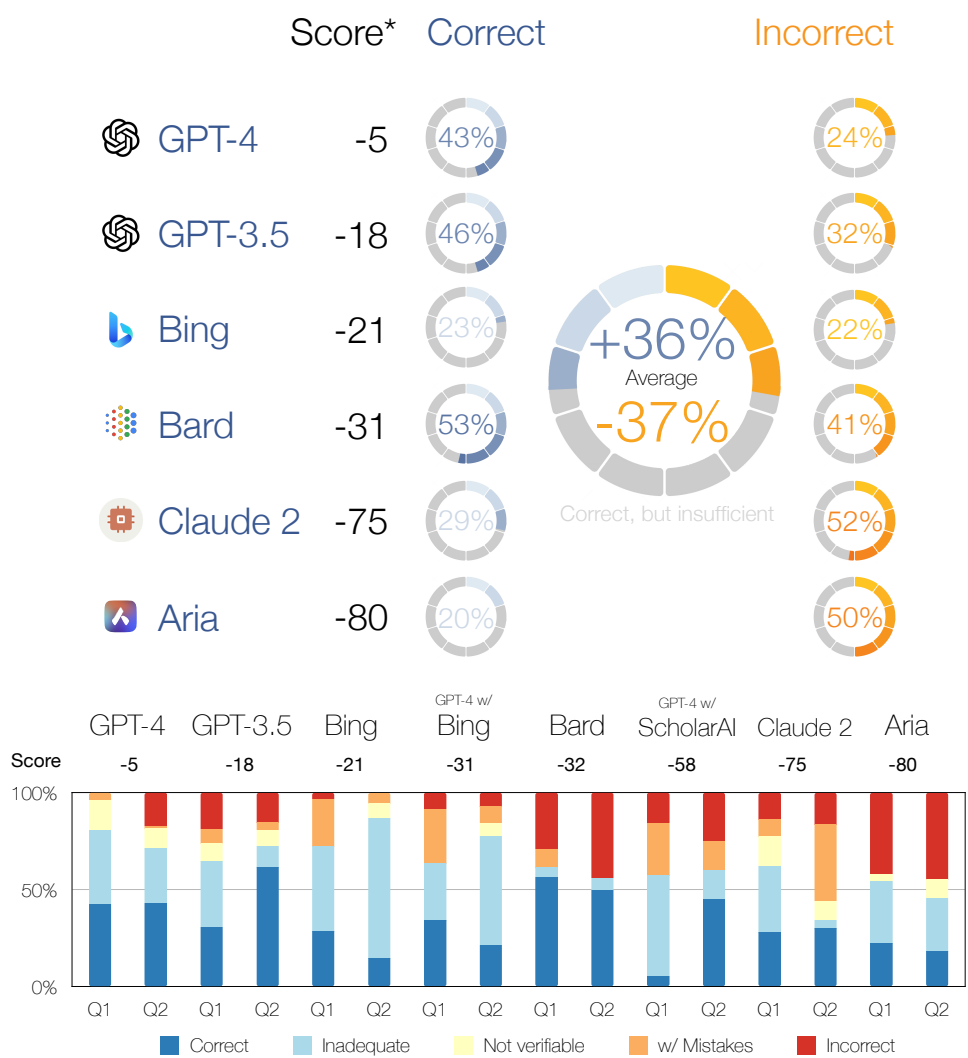

Fig. 4. Quantitative test results, generalized accuracy score above and detailed quantitative accuracy below.

The performance of ChatGPT-3.5 was in line with expectations. On average, it produced approximately the same amount of correct content as ChatGPT-4, but almost 10% more incorrect content. The latter is the expected result of continuous reinforcement learning from human feedback [75], but the former is somewhat underwhelming given the 10-fold increase in the number of parameters in the model.

The Bing chatbot generated only 28% and 13% incorrect or unverifiable content for Q1 and Q2 respectively. However, the percentage of correct content was also one of the lowest at 22% and 18%. The reason for this is that the content was almost exclusively derived from the corresponding Wikipedia pages. The Wikipedia source pages were correctly listed under references, but were systematically misquoted in the text. All references except Wikipedia were secondary, i.e. the generated content copied a selection of references listed on the Wikipedia pages without actually consulting them (many of them are copyrighted material not available on the public web). The article directly referring to Q2 [92] was neither used as a source for the content nor referenced, although, as mentioned above, it is returned by the Bing search engine as the highest-ranking non-Wikipedia result. The content generated by the Bing chatbot thus corresponded well to the chosen settings for a professional (i.e., closely following sources) blog (i.e., formatted in a free flowing text), but fell short of simulating an essay by, for example, a senior undergraduate student.

ChatGPT-4 w/Bing performed significantly worse than Bing Chatbot and also worse than ChatGPT-4. In addition, it was unable to source the content for Q2 from the relevant scientific article. At the time of the test, this plugin was not yet working as intended.

Bard was far less stable and reliable than ChatGPTs. It generated a comparable percentage of correct content, but almost twice as much incorrect content.

ScholarAI, Claude 2, and Aria performed at notably lower level.

ScholarAI was obviously not yet ready for production use. It systematically generated incorrect references and was unable to source relevant content. It should be mentioned, though, that online research databases often perform better in scientific fields other than the humanities, such as medicine, chemistry and natural sciences in general. There are several reasons for this, including the fact that the most prestigious form of dissemination in the humanities is books, which are not indexed in (so many) scientific databases and are often not available in open access format.

The poor result was surprising for Aria, because it uses a GPT LLM from OpenAI. We can only assume that it uses a model older than GPT-3.5. Also, unlike Bing Chatbot and ChatGPT w/Bing, it did not consult online resources to generate the content, but rather it linked the generated content to online resources.

Claude 2 results are comparable to Aria. We can only assume, that it is using a scaled down LaMDA LLM tuned not to outperform Bard.

In summary, we can say that GPT-4, Bing Chatbot, and ChatGPT w/ Bing were close to the level of, for example, an undergraduate student's initial research steps for a term paper, which start with looking up general sources like Wikipedia. GPT-3.5 and Bard were a good substitute for searching the internet with general search engines. They were at a level that could be described as a layman looking at a new topic. ScholarAI, Aria, and Claude 2 were not yet up to the task of answering complex humanities questions.

*3.2 Qualitative precision: AI chatbots' take on original scientific contribution*

The main focus of our article was on whether the tested AI chatbots are able to generate original scientific contribution. The short and expected answer is no. A more detailed answer can be found below (Fig. 5; Appendix B).

As mentioned earlier, human-generated scientific articles in the humanities are typically a combination of derivative and original scientific contributions. In our case study, the human-generated content included ½ of the original scientific contribution for Q1 and ¾ for Q2.

The AI chatbots did not reach this level by far. The only discernible original scientific contribution was at 11% generated by the ChatGPT. ChatGPT-4 aptly inferred in Q1 that the migration of the South Slavs was not a singular event (Appendix B: L. 91—93) and its introductory paragraph in Q2 was extremely apt, profound, and on the cutting edge of science (Appendix B: L. 478—483). Similarly, ChatGPT-3.5 summarised the settlement of the Alpine Slavs very astutely, if repetitively (Appendix B: L. 458—461 and 646—467).

Claude 2 correctly pointed out that the fact that Christian missionaries had to preach in Slavic languages proves the demographic dominance of the Slavs. This is an established historical fact, but not commonly referred to in the context of migration, and was therefore tagged as an original scientific contribution.

ScholarAI has generated what at first sight appeared to be very exciting original scientific content. It has established a direct link between the process of settlement of the Alpine Slavs and their cultural practices and beliefs (Appendix B: L. 578—580 and 585—588). The discussion of beliefs in the context of the migrations is far from the norm and, to our knowledge, has never been brought forward for the migrations of the Alpine Slavs. However, ScholarAI's argumentation was flawed because it was based on irrelevant knowledge pertaining to the Baltic Slavs [102] dwelling about 1000 km northeast of the Alpine Slavs. Interestingly, the same hypothesis could have been argued with another freely available scientific text [103], but this is a book rather than an article and is therefore not in the ScholarAI database.

Other AI chatbots have not generate original scientific contributions.

In conclusion, ChatGPT-4 was once again the best among the AI chatbots tested, but not on the same scale as the human-generated content (Fig. 5).

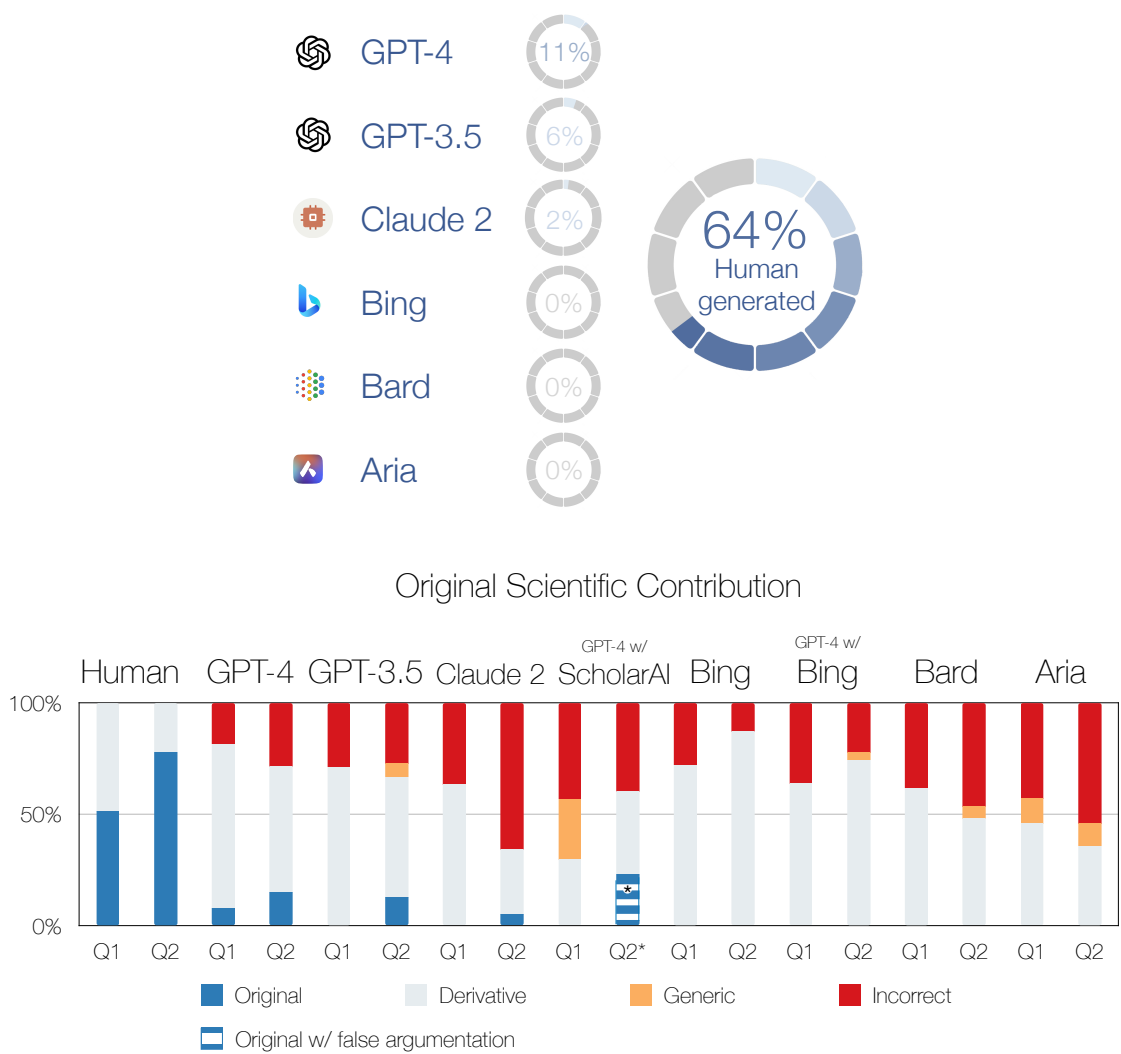

Fig. 5. Qualitative test results, original scientific contribution above and detailed precision below (* false argumentation).

*3.3 Reasoning errors, hallucinations, biases*

The most commonly cited shortcomings of AI chatbots are reasoning errors, hallucinations, and biases, e.g., [9,13]. The terms themselves are not the best choice, because they inappropriately anthropomorphize AI chatbots. However, they are widely used and we have used them for clarity.

In the quantitative analysis above, these shortcomings were interchangeably tagged as 'incorrect,' 'with errors,' or 'unverifiable' (Appendix A). Here we address them qualitatively, on a case-by-case basis.

*Reasoning errors*, also termed lack of on-the-fly reasoning or lack of critical thinking, are the kind of incorrect content in causal statements where cause and effect do not match. Critical thinking is one of the most important qualities for humanities scholars and knowledge workers in general. However, AI chatbots based on LLMs are not designed for this task.

The most obvious example of a reasoning error in our case study was the content generated by ChatGPT-4 and Bard, which causally linked the migration of Slavs into the Alps to the period of climate cooling (Appendix A: L. 512—515 and 703—704). Similarly, ChatGPT-3.5 linked the settlement of the Alpine areas to "fertile lands" (Appendix A: L. 456—459). For most Europeans and most people with formal education worldwide, the Alps are synonymous with mountains and hence with cold climate and harsh agricultural conditions. Most people would therefore reason that a cooling climate and the search for fertile land would not expediate the migration into the Alps, but rather impede it.

Another example of a reasoning error was that almost all tested AI chatbots listed the decline of the (Western) Roman Empire as one of the attractors for the migration of South Slavs to the Balkans (Western Roman Empire: Appendix A, L. 116—117, 135—136, 144—145, 278—281, 322—323, 454—455, 683—684, 736—738; Roman Empire: Appendix A, L. 124—125, 506—507, 516—517, 548—550, 584—585, 593—595). However, we learn in the high school history classes that the (Western) Roman Empire preceded the migration of the South Slavs by at least a century. In fact, the Byzantine Empire was the declining superpower that created the power vacuum for the immigration of the South Slavs to the Balkans.

The fact that both LaMDA (bard) and GPT-4 (ChatGPT-4) generated almost identical incorrect content suggests that such behaviour is inherent in the current generation of GPT LLMs.

The underlying issue on the lack of critical thinking was that none of the tested AI chatbots made any attempt to critically compare different sources. For example, the most important component of a human-generated response to Q1 was: "Currently, there are three main hypotheses..." (Appendix A: L. 394), which was continued by comparing several different sources. No such attempt was detected in the content generated by the AI chatbot. Anecdotally, the majority of randomly selected human users were able to distinguish the critical thinking of the human expert from the content generated by ChatGPT, based solely on the 24-character snippet "There are 3 hypotheses…" without further context (Fig. 6).

Critical comparison of different sources is typical and vital not just in any kind of scientific reasoning, but also in everyday life. The one-sided approach of the tested AI chatbots amplifies "the loudest voice" (the highest ranking search engine result), which is not only bad science but also a grave danger for balanced news reporting, democracy, minority rights, etc.

*Hallucinations* or confabulations of AI chatbots are confident responses by an AI that are not justified by its training data. This is not typical of AI systems in general, but is relatively common in LLMs, as the pre-training is unsupervised [104].

The most obvious hallucinations in our case study were invented references (ChatGPT-4, Appendix C: L. 36; ChatGPT-3.5, Appendix A: L. 495-498; ScholarAI, Appendix A: L. 189—197). Similarly, attempts to inline citations by Bing (Appendix A: L. 226—

257) and ChatGPT-4 w/ Bing (Appendix A: L. 575—581) were largely confabulations. It would appear that the inability to provide correct inline citations is a known problem, as ChatGPT-4 w/ Bing generated a warning to this effect (Appendix A: L. 156—157).

Another very clear example of hallucination was the Late Antique Little Ice Age phenomenon. ChatGPT-4 dated it "between 300 and 700 AD" (Appendix A: L. 512). The correct dates are 536 to about 660 AD, which is clearly intelligible to any human by consulting the title of the reference which ChatGPT-4 correctly provides: "… Late Antique Little Ice Age from 536 to around 660 AD" (Appendix A: L. 532—535).

The underlying issue and a key challenge for current technology, is that "AI chatbots do not know what they do not know" and may very confidently invent facts [46] rather than formulate the sentence "I don't know".

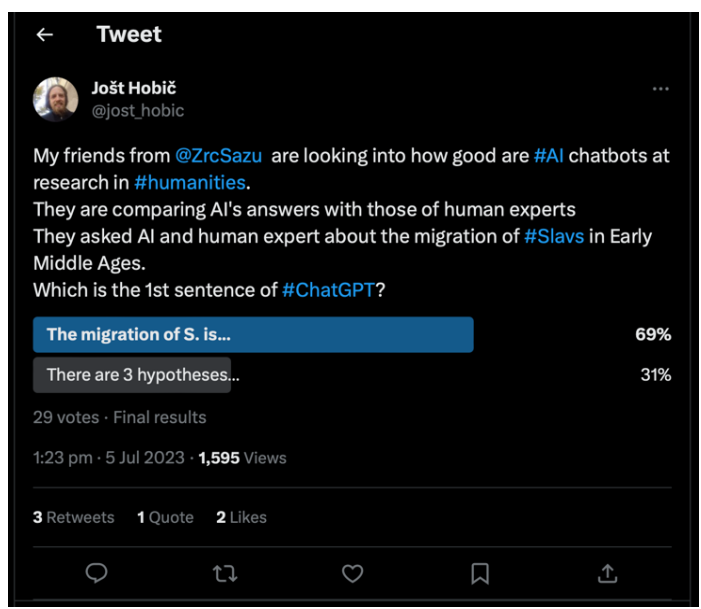

Fig. 6. Twitter (now X) poll asking human users to differentiate between ChatGPT and human-generated content with almost no context. Most respondents answered correctly.

*Biases* are often exhibited by AI chatbots. They are based on training data, but according to recent research, they can be amplified beyond existing perceptions in society. Biases generated by AI chatbots can therefore be informative about the underlying data, but they can also be misleading if the AI-generated content is used uncritically. The most researched biases to date are those related to gender, race, ethnicity, and disability status, e.g., [29,49,104–106].

In our test we have detected three different types of biases: language bias, neo-colonial bias, and citation bias.

First, language bias. Although there is far more relevant scholarly content written in Balkan languages than in English, 92% of the references generated by the AI chatbots in our test referred to English and none to Balkan-language publications (Fig. 7). This can only be partially explained by the fact that the prompts were in English. Namely, three (8%) German references prove that English was not the only criterion for selection. When question Q2 was asked with a prompt in Slovenian, two references were again in English and the third in Slovenian was a hallucination (Appendix C: L. 36).

The detected language bias is most likely due in large part to the language bias of online search engine ranking algorithms that favour English publications [107]. This bias seems to be a wasted opportunity, because all tested AI chatbots "understand" many languages, e.g., [13].

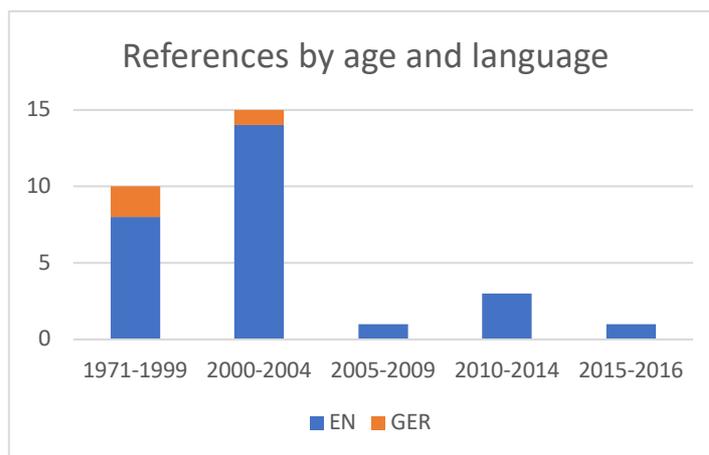

Fig. 7. Language of references generated by AI chatbots to an English language prompt.

Second, neo-colonial bias. 88% of the references are by authors from the global West. Only a minority (12%) belong to English translations of texts originally written in Slavic languages [102,108,109], although there are several other (more) relevant translations, e.g., [88,110–117]. This reflects a scholarly hierarchy created by colonialism (until the 1910s, the Balkans were largely divided between the Austro-Hungarian and Ottoman Empires), sometimes referred to as a neo-colonial pattern in the global network of science in which the intellectual dominance of the global West is growing [118,119]. To our knowledge, the neo-colonial bias for the study of medieval Slavs has not yet been explicitly analysed or even discovered, as it has never been revealed as clearly as through the use of AI chatbots in this case study.

Third, citation bias. 75% of the references are from before 2005 and the oldest was originally published in 1895 [108]. This is showing a very clear bias towards new and up to date publications. For example, by far the most referenced publication in our case study is Curta [85]. While this is still a seminal work, it is outdated and has often been criticised, e.g., [91] and the critiques have been responded to [86]. Therefore, in a modern scientific text created by a human expert, the reference to Curta is always followed by either its critique or an up-to-date response to that critique. In AI-generated content, however, Curta is always referenced as the primary source.

This bias is in line with the growing trend to cite old documents, caused at least in part by the "first page results syndrome" combined with the fact that all search engines favour the most cited documents [120,121]. The ScholarAI plugin, for example, transparently discloses that it ranks references solely by the number of citations (Appendix A: L. 622, footnote 25).

These findings are consistent with the recent study looking at references generated by ChatGPT-3.5. It revealed that ChatGPT-3.5 tends to reference highly cited publications, shows a preference for older publications, and refers predominantly to reputable journals. The study concluded that ChatGPT-3.5 appears to rely exclusively on Google Scholar as a source of scholarly references [56].

All three of the listed biases are abundant in the training material, which is of course mostly the public web. In other words, an uncritical online research by a human using one of the major search engines, Google Scholar or Wikipedia pages would yield comparable results with language bias, citation bias, and neo-colonial bias. However, proper research by a human expert(s) would avoid citation bias and at least reduce language and neo-colonial bias.

*3.4. Race for parameters*

It was not the intention of this article to gain insights into the efficiency of LLMs, but as a side note, the current trend of upsizing the LLMs, sometimes referred to as an AI Arms Race, can be addressed. Currently, numerous well-funded startups, including Anthropic, AI21, Cohere, and Character.AI, are putting enormous resources into developing ever larger algorithms and the number of LLM parameters exceeds the exponential growth trend line (Fig. 8). However, it is expected that the returns on scaling up the model size will diminish [13].

In our data, we observed the impact of the exponential growth of the LLM on the content generated. ChatGPT-4 is approximately 10 times larger than ChatGPT-3.5. LaMDA has a similar number of parameters as ChatGPT-3.5, 137 and 175 billion respectively, but the AI chatbot tested, Bard, only uses the "lightweight" version of LaMDA. Assuming that the "lightweight" version uses only half the parameters, the size ratio between LaMDA (Bard), ChatGPT-3.5, and ChatGPT-4 is 1:2:20.

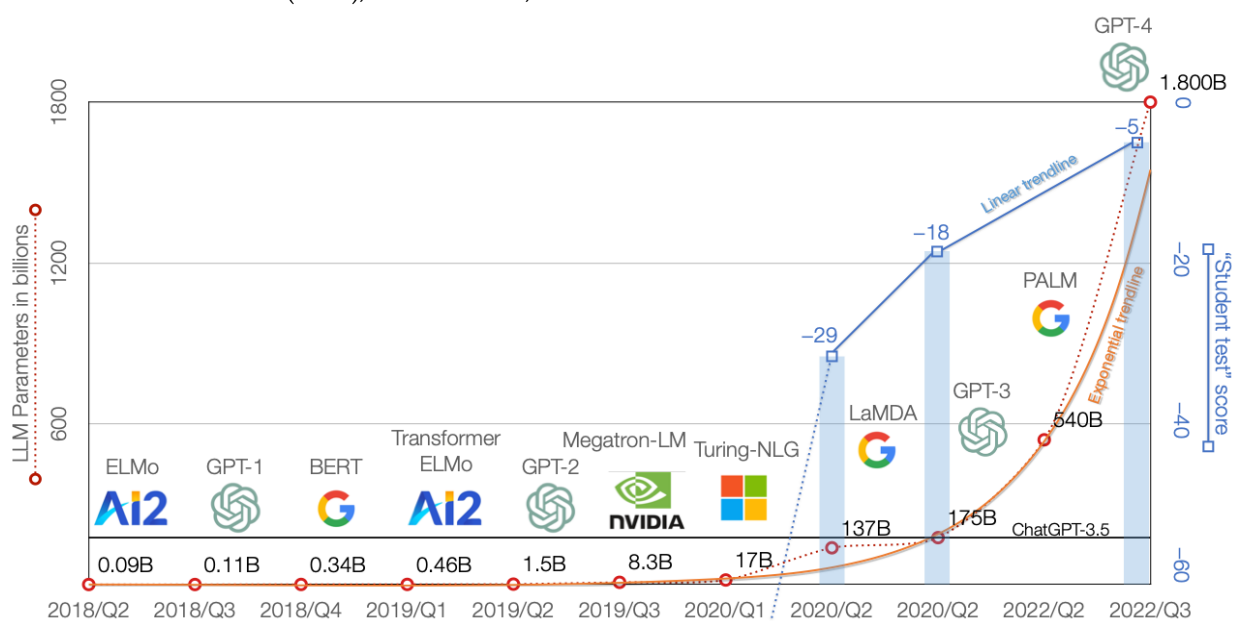

Fig. 8. The race for parameters. The increase in the number of parameters (red dotted line) exceeds the exponential trend (orange line); the in-context learning as detected by our test (blue columns; see section 3.1) only improves with a linear trend (blue line; adapted after [122] and updated with sources cited in section 2.1).

The up to 10% improvement in generated content of ChatGPT-4 compared to ChatGPT-3.5 is not negligible, but for many use cases it might not be noticeable. Given the tenfold increase in the number of parameters, this is not particularly impressive. Especially since the 10% improvement is not solely due to the new model, but also to continuous reinforcement learning from human feedback. Nevertheless, the improvement of ChatGPT-4 over Bard in reducing incorrect content is significant. One can surmise that the twenty-fold increase in parameters from Bard to ChatGPT-4 is significant enough to be noticeable in daily use, while the ten-fold increase from ChatGPT-3.5 to ChatGPT-4 is mostly only observable in a controlled test environment.

Given that improvements are diminishing and that continuous exponential growth of LLMs is physically impossible with current technology, we can only conclude that the growth of LLMs has reached a plateau. This observation is consistent with the sparse information coming out of the industry: "…we're at the end of the era where it's going to be these, like, giant, giant models… We'll make them better in other ways." (Sam Altman, CEO of OpenAI, at an event on MIT on 14 April 2023).

## 4. Discussion: Data to Knowledge

The results of our analysis show that ChatGPT, while currently the most advanced AI chatbot, is not capable of producing original scientific content. This is a zero-shot learning problem, but it is also much more than that. Namely, LLM-based AI chatbots are not designed to generate content in the same way that human researchers produce new knowledge.

A typical process for producing original scientific contribution, which was also used in our case study, is an intricate process that is perhaps best explained using Ackoff's Data-Information-Knowledge-Wisdom (DIKW) hierarchy. According to this hierarchy, data is the raw input, simple facts without context; information is data that has been given context and is meaningful in some way; knowledge is the understanding and interpretation of that information gained through experience or learning; and wisdom, the highest level of the pyramid, is the ability to apply knowledge judiciously and appropriately in different contexts, understand the broader implications, and make insightful decisions based on accumulated knowledge and experience [123].

In our case study, the data was the documentation of the excavations of 1106 archaeological sites. The information was summaries of the excavation reports and scientific publications on these 1106 sites curated in the structured database Zbiva, available on the public web since 2016 [124]. Knowledge were the scholarly articles discussing and/or analysing the migration of the Alpine Slavs, e.g., [97,114,115]. Wisdom is what happens (rarely) after the publication of the scientific articles and after the generation of AI chatbot content and does not concern us here.

Human researchers approached Q2 by first obtaining and refining the data, which was followed by analysing the information using appropriate software, formulating knowledge based on the results, and finally disseminating the knowledge in the form of an original scientific article.

In the real world, archaeological practice (and most humanities research) is messy and therefore this process is fuzzy and recursive. It takes the form of a hermeneutic spiral [125], which from the perspective of computational algorithms are loops between data, information and knowledge. These loops involve solving computationally irreducible processes that cannot be handled by LLM-based AI chatbots alone (Fig. 1: I).

In other words, to generate original scientific content requires not only access to curated data/information, but also the ability to analyse it.

LLMs, on the other hand, are pre-trained on existing knowledge (texts, books) and only able to recombine it in new permutations. Occasionally this can lead to modest original scientific content. For instance, AI chatbot could be used to summarize what a historical text verbatim says about a certain subject, but not to interpret it. Therefore, this is a limited and finite avenue to original scientific contributions. Regardless of the future improvement of LLMs, LLM-based AI chatbots will not be able to replicate the workflow of human researchers, as they are only trained on the existing knowledge.

A purpose-built LLM-based software, on the other hand, could handle such a workflow: searching for data/information, performing relevant analysis, generate textual and graphical information, and summarising it into new knowledge (Fig. 1: K, L, M; G, J). Such LLM-based software would have several qualities of an AGI and is in fact already feasible with existing technology, for example by combining Prometheus, a relevant cloud-based software connected through a ChatGPT API, and ChatGPT-4.

In a nutshell, LLM-based AI chatbots are not and probably will never be able to generate new knowledge from data in the same way as human researchers (in the humanities), but appropriate LLM-powered software might be.

## 5. Conclusion: fluent but not factual

Most commentators on generative AI, including the authors of this article, agree that the current generation of AI chatbots represent AI's inflection point. It is very likely that historiography will record ChatGPT-3 as the eureka moment that ushered in a new era for humanity. What this new era will bring is currently unknown, but it seems that it will, for better or worse and probably both, change the world. To maximize the positive and to mitigate the negative as much as possible, making AI safe and fair is necessary. And a significant part of making AI safe is testing.

The aim of this article was to test the current AI chatbots from human(ities) perspective, specifically their scientific writing abilities. We compared six AI chatbots: ChatGPT-3.5, ChatGPT-4, Bard, Bing Chatbot, Aria, and Claude 2.

In accordance with expectations, ChatGPT-4 was the most capable among the AI chatbots tested. In our quantitative test, we used a method similar to grading undergraduate students. Bing Chatbot and ChatGPT-4 were nearing the passing grade and ChatGPT-3.5 and Bard were not far behind. Claude 2 and Aria produced much weaker results. The ChatGPT-4 plugins were not yet up to the task.

In our qualitative test, we used a method similar to peer reviewing a scientific article. ChatGPT-4 was again the best performer, but it didn't generate any notable original scientific content.

Additional shortcomings of the AI-generated content that we found in our test include reasoning errors, hallucinations, and biases. Reasoning errors refer to the chatbot's inability to critically evaluate and link cause-and-effect relationships, as evidenced by several historical inaccuracies regarding the migration patterns and historical periods of the Slavs. Hallucinations denote confident but unsubstantiated claims by the AI, such as invented references and inaccurate dates. Our test also reveals significant biases in the content generated by the AI. These biases manifest as language bias, favouring English sources over other relevant languages; neo-colonial bias, displaying a preference for Western authors; and citation bias, skewing towards older and more highly cited publications. These findings highlight that despite their technological prowess, AI chatbots remain reliant on their training data echoing or even amplifying existing biases and knowledge gaps. Because they veer towards past data, they are likely to be too conservative in their recommendations. Since the listed deficiencies are relatively inconspicuous compared to, for example, gender or racial biases, it is unlikely that they will be remedied in the foreseeable future by resource-intensive reinforcement learning from human feedback. These biases are among the key concerns with the use of AI chatbots in scientific writing, as they are less likely to be highlighted in the review processes.

Our results also highlighted the possible future trends in the development of AI chatbots. The large discrepancy between almost passing an undergraduate exam and not producing any notable scientific contribution may seem surprising at the first glance. On closer inspection, however, this was to be expected. "Doing science", i.e. making an original scientific contribution, is much more complex than just doing very well in the exams. It is based on proactively acquiring and analysing data and information to generate new knowledge, whereas "passing an exam" is based on accumulating existing knowledge and passing it on at a request.

"Passing an exam" will further improve when AI chatbots are given access to curated data. An AI chatbot with access to selected datasets would be a typical downstream task developed around an existing LLM. However, without access to external tools, LLM-based AI chatbots will never be suitable for "doing science". Therefore, in the near future an evolution of current LLM-based AI chatbots towards LLM-powered software capable of, among other things, "doing science" seems likely. This assertion is in line both with the fact that the growth of LLMs seems to have plateaued and that the industry is turning to other solutions.

In conclusion, we agree with the previous commentators that AI chatbots are already widely used for various tasks in scientific writing due to their applicability to a wide range of tasks. However, AI chatbots are not capable of generating a full scientific article that would make a notable scientific contribution to the humanities or, we suspect, to science in general. If left unsupervised, AI chatbots generate content that is fluent but not factual, meaning that the errors are not only many but also easily overlooked, *cf.*[126]. Therefore, peer review processes need to be rapidly adapted to compensate for this, and the academic community needs to establish clear rules for the use of AI-generated content in scientific writing. The discussion about what is acceptable and what is not must be based on objective data. Articles like this one are necessary to support those decisions and we suspect that many more will follow.

As for the future beyond the immediate foreseeable, when LLM-powered software and/or AGI will be able to generate original scientific contributions, we agree that questions about explainable AI are likely to come to the fore. Understanding our world, a fundamental aspiration of the humanities, will only be partially achieved through the use of black-box AI. Since the humanities, just like justice, for example, are as much about process as outcome, humanities scholars are unlikely to settle for uninterpretable AI-generated predictions. We will want human-interpretable understanding, which is likely to be the remaining task of human researchers in the humanities in the future, *cf.* [127].

**Supplementary Materials:** Appendix A, Appendix B, and appendix C are available on the open access repository Zenoto under CC-BY 4.0 lince: https://doi.org/10.5281/zenodo.8345088.

**Author Contributions:** Both authors, E.L. and B.Š., contributed equally to the article. Conceptualization, E.L. and B.Š.; methodology, E.L. and B.Š.; validation, E.L. and B.Š.; formal analysis, E.L. and B.Š.; writing—original draft preparation, E.L. and B.Š.; writing—review and editing, E.L. and B.Š.; visualization, E.L. and B.Š.; project administration, E.L. and B.Š.; funding acquisition, E.L. and B.Š. All authors have read and agreed to the published version of the manuscript.

**Funding:** This research was part of the AI4Europe project that has received funding from the European Union's Horizon Europe research and innovation programme under Grant Agreement nº 101070000.

**Institutional Review Board Statement:** Not applicable.

**Informed Consent Statement:** Not applicable.

**Data Availability Statement:** All data used in and produced by the research are available in the appendices.

**Acknowledgments:** The authors give thanks to dr. Zoran Čučković for introducing them to ChatGPT in December 2022.

**Conflicts of Interest:** The authors declare no conflict of interest. The funders had no role in the design of the study, in the collection, analyses, or interpretation of data, in the writing of the manuscript, or in the decision to publish the results.